\documentclass[runningheads,a4paper]{llncs}
\usepackage{siunitx}

\def\etal{\textit{et al.}\xspace}

\usepackage{xspace}
\usepackage{amssymb}

\usepackage{graphicx}
\usepackage{amsmath}

\usepackage{algorithm}
\usepackage[noend]{algpseudocode}

\usepackage{array}
\usepackage{color, colortbl}

\makeatletter
\renewcommand{\ALG@beginalgorithmic}{\small}
\makeatother

\usepackage{url}

\definecolor{lightgray}{gray}{0.95}

\begin{document}

\mainmatter  % start of an individual contribution

%\title{Multiple Organ Point Cloud Network for prediction on Medical Shape Data}
%\title{Towards the use of Deep Learning shape descriptors. }
\title{Deep Shape Analysis on Abdominal Organs \\for Diabetes Prediction}
\titlerunning{}

%
%\author{No Author given}

\author{Benjam\'in Guti\'errez-Becker \inst{1} \and Sergios Gatidis \inst{2}\and Daniel Gutmann \inst{2} \and Annette Peters  \inst{3} \and Christopher Schlett \inst{4} Fabian Bamberg \inst{2} \and  Christian Wachinger \inst{1}} 
\authorrunning{Benjam\'in Guti\'errez-Becker \etal} 
\institute{
     Artificial Intelligence in Medical Imaging (AI-Med), KJP, LMU M\"unchen\\
      \and
      Department of Diagnostic and Interventional Radiology, University of T\"ubingen\\
      \and 
      Institute of Epidemiology, Helmholtz Zentrum M\"unchen
      \and 
      Department of Diagnostic and Interventional Radiology, University Hospital Heidelberg}

\toctitle{Lecture Notes in Computer Science}
\tocauthor{Authors' Instructions}
\maketitle
%\footnotetext{Accepted at MICCAI 2018.}
\begin{abstract}
Morphological analysis of organs based on images is a key task in  medical imaging computing. Several approaches have been proposed for the quantitative assessment of morphological changes, and they have been widely used for the analysis of the effects of aging, disease and other factors in organ morphology.
In this work, we propose a deep neural network for predicting diabetes on abdominal shapes. The network directly operates on raw point clouds without requiring mesh processing or shape alignment. 
Instead of relying on hand-crafted shape descriptors, an optimal representation is learned in the end-to-end training stage of the network. For comparison, we extend the state-of-the-art shape descriptor BrainPrint to the AbdomenPrint. Our results demonstrate that the network learns shape representations that better separates healthy and diabetic individuals than traditional representations. 
\end{abstract}

\section{Introduction}

%The study of the morphology of anatomical structures is an important task in medical imaging. 
%The use of shape representations for the morphological analysis of anatomical structures is an important task in medical imaging. 
Shape models have been widely used in medical imaging, not only as a prior for segmentation algorithms, but also as a powerful tool to  assess morphological differences between subjects \cite{heimann2009statistical}.  A critical element in shape analysis  is the choice of a numerical representation which can be used for a quantitative analysis of shape. Multiple shape representations have been previously explored, ranging from very basic volumetric and thickness measurements \cite{valizadeh2017age,becker2018gaussian}, to more complex models such as  Point Distribution Models \cite{Cootes1995}, spectral signatures \cite{wachinger2015brainprint}, spherical harmonics \cite{gerardin2009multidimensional}, medial representations \cite{Gorczowski2007}, and diffeomorphisms \cite{miller2014diffeomorphometry}. % or learned representations \cite{Shakeri2016,gutierrez18shape}. 

Despite the ample success of deep learning for many medical imaging tasks, their application for medical shape analysis is still largely unexplored; mainly because typical shape representations such as point clouds and meshes do not possess an underlying Euclidean or grid-like structure. Deep networks can learn complex, hierarchical feature representations from data that typically outperforms hand-crafted features, which are not optimal for the given task.

Recently, we have introduced the Multi-Structure PointNet (MSPNet) \cite{gutierrez18shape}, which is able to learn shape representations directly on point clouds and can predict a label given the shape of multiple brain structures. To the best of our knowledge MSPNet is the first deep end-to-end learning system used to perform prediction based on organ shapes. MSPNet operates directly on point clouds, without the need to create meshes and it does not require  computing point correspondences between different shapes. 

As most work on shape analysis, we have used MSPNet in the study brain morphology; however,  the use of shape models to analyze other anatomical regions remains  a relatively unexplored area. In an effort to fill this gap, we propose the deep shape analysis of abdominal anatomy. Our main interest lies on the use of MSPNet to learn shape representations which are able to measure morphological differences in the liver and spleen of healthy subjects when compared to  individuals diagnosed with diabetes mellitus. Diabetes mellitus is a worldwide prevalent condition, which is defined by levels of hyperglycaemia giving rise to risk of microvascular damage and its diagnosis is associated with complications, which lead to reduced life expectancy and diminished quality of life \cite{world1999definition}. 

Concretely, we propose the first deep learning approach operating on the shape of  abdominal organs for the prediction of diabetes. Further, we extend the state-of-the-art shape representation BrainPrint \cite{wachinger2015brainprint} to the abdomen, yielding the AbdomenPrint. Finally, we compare MSPNet and AbdomenPrint in the challenging task of predicting  diabetes  directly from the shape of the liver and spleen.

%Despite the wide availability of shape representations, their use  has mainly focused on the analysis of brain morphology and their use in the analysis of other body organs has been relatively unexplored.  In an effort to fill this gap, in this work, we propose the use of two state of the art  shape models for the analysis of abdominal anatomy. First, we propose AbdominalPrint, which corresponds to the use of ShapeDNA features \cite{wachinger2015brainprint} for the analysis of the morphology of abdominal organs. Second, we propose the use of MSPNet \cite{gutierrez18shape}: a deep learning based solution operating directly on point cloud representations.  Our main interest lies on the use of these shape representations to measure morphological differences in the liver and spleen of healthy subjects when compared to  individuals diagnosed with diabetes mellitus. Diabetes mellitus is a worldwide prevalent condition which is defined by levels of hyperglycaemia giving rise to risk of microvascular damage and its diagnosis is associated with complications which lead to reduced life expectancy and diminished quality of life \cite{world1999definition}.

\subsection{Related Work}
\def\real{{\mathbb{R}}}
\def\ourmethod{{MSPNet}}

The use of shape models for the analysis of morphological changes associated with disease or other factors has mainly been explored in neuroimaging. Significant relationships between measurements of brain morphology and a variety of factors such as age \cite{cole2017predicting} and neurodegenerative diseases \cite{wachinger2015brainprint} have been throughly explored. In this work, we focus on the abdominal organs liver and spleen.  Previous approaches have explored the morphological analysis of the liver based on imaging data. Lamecker \etal  \cite{lamecker2004segmentation} present for the first time a statistical shape model of the liver. Dura \etal \cite{dura2017probabilistic} present the construction of a a probabilistic liver atlas. In terms of using shape models for the diagnosis of liver related diseases, Kohara \etal \cite{Kohara2010} use a statistical shape model of the liver to assess differences between healthy subjects and individuals diagnosed with cirrhosis. A similar approach is proposed by Mukherjee \etal \cite{Mukherjee2013} for the discrimination of chronic liver disease from CT Data and by Hori \etal \cite{Hori2015} where a statistical shape model is used to evaluate differences in liver shape caused by hepatic fibrosis. 

Shape analysis of the spleen is a far less explored area of research. Tateyama \cite{Tateyama2009} \etal present the use of a Point Distribution Model (PDM) for the analysis of spleen shape. Yates \etal \cite{Yates2016} present a morphological study of the spleen, relating the principal components obtained from a statistical shape model to anthropometric and demographic information. 

In this work, we deviate from these previous approaches in the methodology used to assess relationships between abdominal morphology and clinical variables. Instead of modeling shape variation using the commonly used features derived from Point Distribution Models, we evaluate the use of two  state-of-the-art approaches: MSPNet\cite{gutierrez18shape} and BrainPrint \cite{wachinger2015brainprint}. These two approaches have previously been used for the morphological analysis of brain structures, but they have yet to be applied for the analysis of abdominal structures.

\section{Method}
\noindent
The usual pipeline of shape analysis of anatomical structures consists of extracting a binary segmentation of the structure of interest from an image (either manually or automatically) followed by the extraction of a shape descriptor vector $\mathbf{S}\in\real^{dim}$ which can be used to quantitatively model the shape of an organ of interest.  In the case of a classification task,  we can then find a function $f(\mathbf{S})\mapsto y$ mapping  shape descriptors $\mathbf{S}$ to a label $y$. which corresponds to the variable to be predicted. In our case $y\in \{0,1\}$ is an indicator variable which determines if a particular subject is healthy $(y=0)$ or has been diagnosed with a diabetic condition $(y=1)$. 

In our experiments, we evaluate the use of two different shape representations $\mathbf{S}$ . In both cases, independent shape representations $\mathbf{S}_l$ and $\mathbf{S}_s$ are calculated independently for the liver and spleen, and are afterwards concatenated to obtain a global shape descriptor $\mathbf{S}$.

\begin{figure}[t]
\centering\includegraphics[width=0.9\columnwidth]{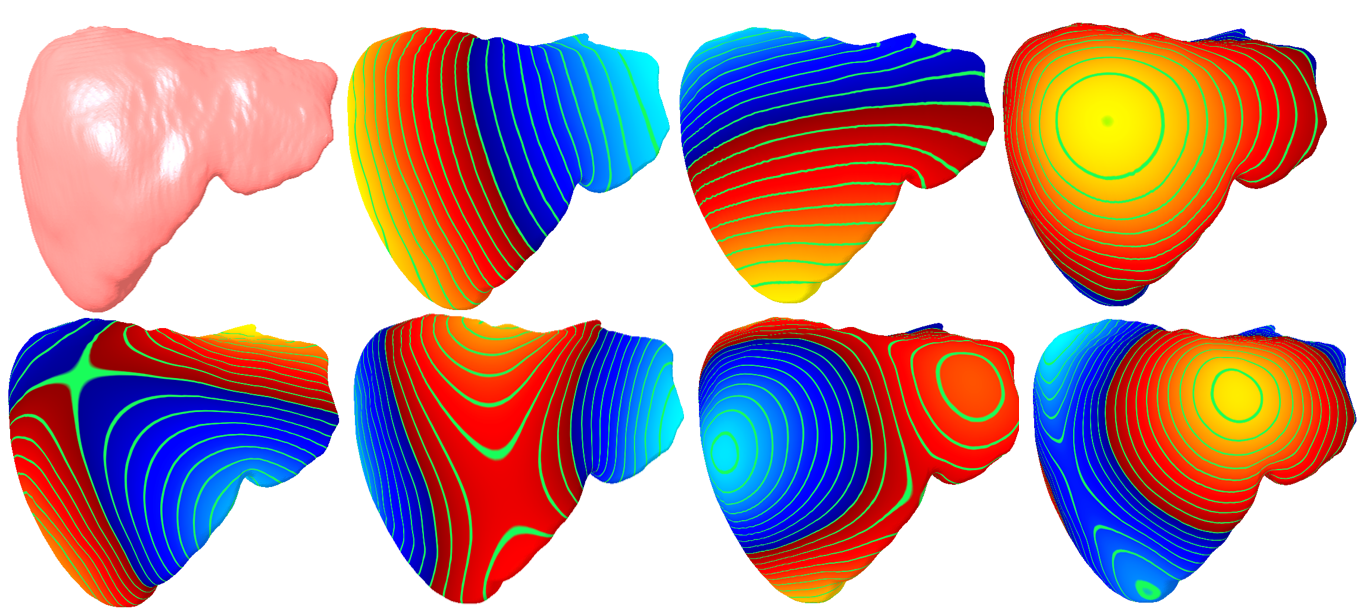}
\caption{Liver surface and first seven non-constant eigenfunctions of the Laplace-Beltrami operator (sorted left to right, top to bottom) calculated on the surface. Increasing positive values of the eigenfunctions are shown in the color gradient from red to yellow and decreasing negative values are shown from dark blue to light blue. }
\label{fig:liver_efs}
\end{figure}

\subsection{AbdomenPrint} 
The AbdomenPrint $\mathbf{S}$ is the analogy of the BrainPrint \cite{wachinger2015brainprint}, which has been successfully used to associate morphological changes in the brain correlated to  Alzheimer's disease \cite{wachinger2016domain,wachinger2016whole,wachinger2018longitudinal}, but in our case we apply it to the analysis of  abdominal organs. AbdomenPrint uses the shapeDNA \cite{reuter2006laplace} as shape descriptor, which is computed from the intrinsic geometry of organs by calculating the Laplace-Beltrami spectrum. Considering the Laplace-Beltrami operator $\Delta$, the spectrum is obtained by solving the Laplacian eigenvalue problem:

\begin{equation}
\Delta f = -\lambda f.
\end{equation}

The solution of this problem consists of a series of eigenvalues $\lambda_i \in \real$ and eigenfunctions $f_i$ (see figure \ref{fig:liver_efs}). The first $l$ non-zero eigenvalues, computed with the finite element method, form the ShapeDNA: $\mathbf{\lambda} = (\lambda_1, \dots, \lambda_l)$.   
We further linearly re-weight the eigenvalues, $\hat{\lambda}_i = \lambda_i / i$, to balance the impact of higher eigenvalues that show higher variance~\cite{wachinger2015brainprint}. The shape of an organ can then be represented by the vector of normalized eigenvalues $\mathbf{S} = \mathbf{\hat{\lambda}}$.  For the computation of shapeDNA, triangular meshes are constructed from organ segmentations via marching cubes.

\subsection{MSPNet}  
We have recently introduced Multi-structure PointNet (MSPNet) \cite{gutierrez18shape}  for shape analysis of brain structures. MSPNet  is a network architecture based on PointNet, a state of the art deep learning approach for point cloud classification  \cite{Qi2017}.  In MSPNet, a shape representation can be learned in an end-to-end fashion directly from a point cloud $ \mathbf{P} = [x_0,y_0,z_0, x_1,y_1,z_1,...,x_n,y_n,z_n] $ where  $x_i, y_i,z_i$ correspond to the cartesian coordinates of the points representing the surface of the organ of interest. Different to other shape representations based on point clouds such as Point Distribution Models, in MSPNet it is not required for  the points in $\mathbf{P}$ to be ordered, which means that no anatomical correspondences between shapes are needed.

\begin{figure}[h]
\centering\includegraphics[width=0.9\columnwidth]{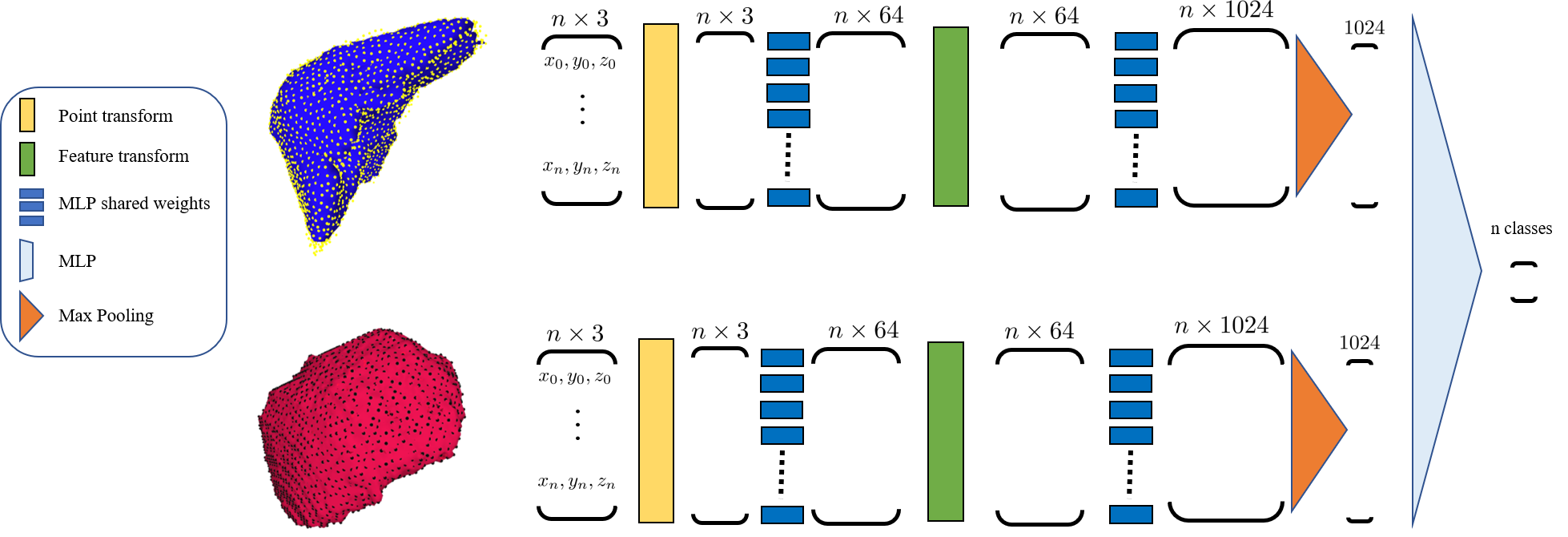}
\caption{MSPNet Architecture for abdominal structures. The network consists of one branch per structure, one for the liver and one for the spleen, which are fused before the final  multilayer perceptron (MLP). Each structure is represented by a point cloud with $n$ points that pass through transformer networks and multilayer perceptrons at  each individual branch. Numbers over each layer correspond to their sizes.}
\label{fig:network}
\end{figure}

To obtain  a shape representation using MSPNet, the point cloud vector $\mathbf{P}$ is fed to the network (see fig. \ref{fig:network}). The first stage of the network corresponds to a transformation network which corresponds to a function $f(\mathbf{P}) \to \mathbf{T}$  mapping the input point cloud to a transformation matrix $\mathbf{T} \in \real^{3x3}$ . This transformation matrix is applied to the input point cloud, so that the input point clouds are aligned before further processing is done. This transformation layer is known as T-Net \cite{Qi2017}, and is similar in structure to PointNet. After this transformation is applied to the input point cloud, the representation $\mathbf{S}$ is obtained by applying $l=1...L$ layers:

\begin{equation}
\mathbf{S}^{(l)} =\mathbf{max}(\{ g( W^{(l)} a_i^{(l-1)} + b^{(l)} ) \}_{i=1}^N )
\end{equation}
where $W^{(l)}$ corresponds to the shared weights of the $l$th layer, $g$ is a non linear activation function and $a_i^{(l-1)}$ correspond to the activation of the $i$th point of the previous layer. By processing the point cloud $\mathbf{P}$ through these shared weight layers, MSPNet obtains a global feature vector  $\mathbf{S}$ at the last shared weight layer corresponds to a feature representation of each organ.  This feature vector is then connected to a fully connected multilayer perceptron corresponding to a function $f(\mathbf{S})\mapsto y$. It is important to notice that since this network is trained in an end-to-end fashion, the  feature vector $\mathbf{S}$ is optimized for the diabetes classification task.

\section{Experiments}
Experiments are performed on a set of whole-body Magnetic Resonance Images (MRI) obtained from the Cooperative Health Research in the Region Augsburg project (KORA). Manual segmentations of the liver and the spleen were obtained from 359 images, 228 corresponding to healthy controls and 131 corresponding to subjects  diagnosed with either pre-diabetes or diabetes according to definitions by the world health organization \cite{world1999definition}.  From these segmentations, point clouds  are obtained by uniformly sampling the surface area of each organ. 
%From these segmentations, triangular meshes of the liver and spleen are obtained using the marching cubes algorithm as implemented in the Visualization Toolkit (VTK). For 

%\begin{figure*}[h!]
%	\begin{minipage}[b]{0.50\linewidth}
%		\centering\includegraphics[width=\textwidth]{figures/roc_pointnet}
%		\label{fig:roc_pointnet}
%	\end{minipage}
%	\hfill
%	\begin{minipage}[b]{0.50\linewidth}
%		\centering\includegraphics[width=\textwidth]{figures/roc_shapeDNA}
% \label{fig:roc_shapeDNA}
%	\end{minipage} \\
%	\caption{ROC curves for the classification of healthy and individuals diagnosed with pre-diabetes or diabetes. }\label{fig:rocs}
%\end{figure*}
\begin{figure}[h]
\centering\includegraphics[width=0.9\columnwidth]{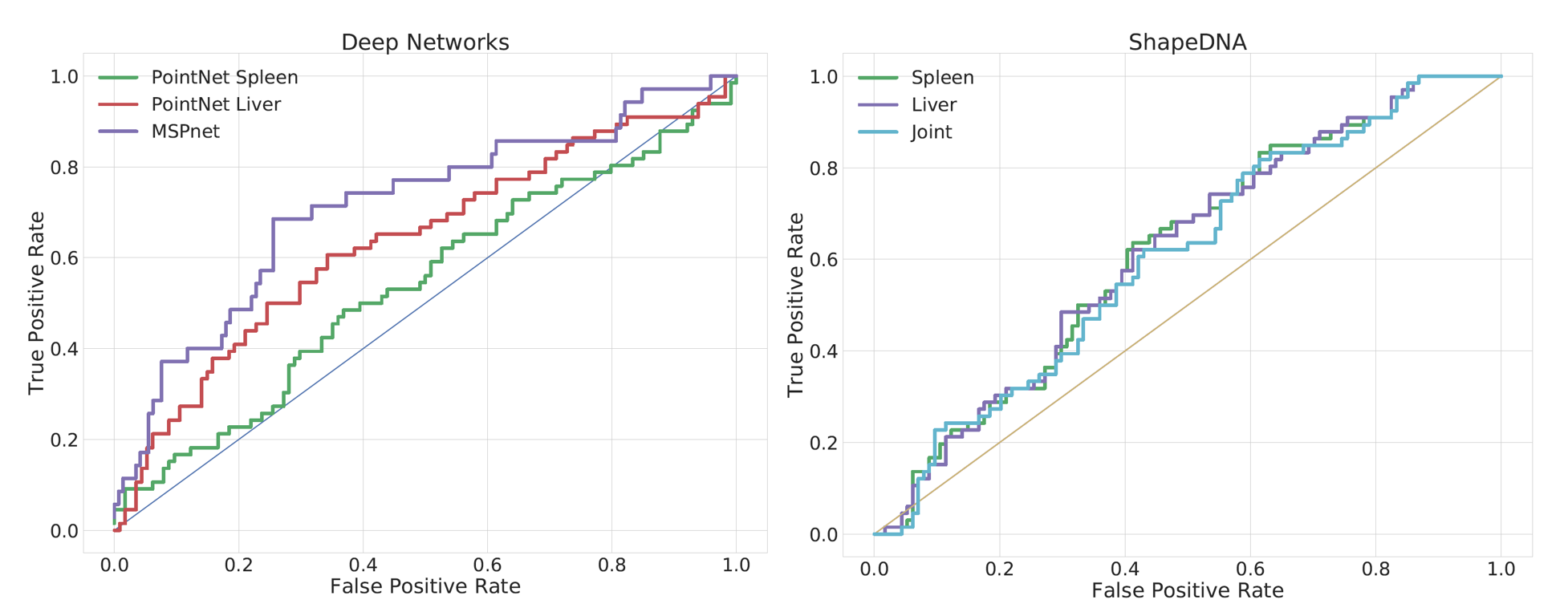}
\caption{ROC curves showing the classification results between healthy subjects and individuals diagnosed with either pre-diabetes or diabetes. Higher area under the curve values were obtained for MSPNet operating on spleen and liver shapes simultaneously.}
\label{fig:rocs}
\end{figure}

\subsection{Diabetes classification}
For a first experiment we evaluate the ability of each shape representation for the problem of discriminating between shapes of organs obtained from healthy individuals compared to subjects diagnosed with pre-diabetes or diabetes. Our set of 359 images is split divided in training and testing sets (50/50) and classification performance is evaluated in terms of Area Under the Curve (AUC) (Table \ref{table:classification}) of the Receiver Operating Characteristic (ROC) curves shown in figure \ref{fig:rocs}.  For comparison we compute AbdomenPrint features, and a gradient boosting classifier is trained to operate on the obtained shape descriptors.
Our results show that although both methods are able to detect differences between diabetic and control patients, MSPNet presents a higher classification performance. It is also worth mentioning that using joint shape descriptors of both the spleen and the liver did not improve classification for AbdomenPrint, whereas MSPNet was able to leverage on joint information obtained from both organs simultaneously.

 \begin{table*}
 	\centering
  	\caption{Area under the curve values for the diabetes classification experiment.}
 		\begin{tabular}{ l c c c }
 			  	&  Liver &  Spleen &   Abdomen  \\ 
               \rowcolor{lightgray} 
              AbdomenPrint 	&0.61 & 0.60 & 0.62 \\
               MSPNet & 0.69 &  0.61 & \textbf{ 0.74}   \\ 
 			\end{tabular}
 	\label{table:classification}
 \end{table*}

\subsection{Visualization of the Shape Feature Spaces}

\begin{figure*}[h!]
  		\centering\includegraphics[width=1.0\textwidth]{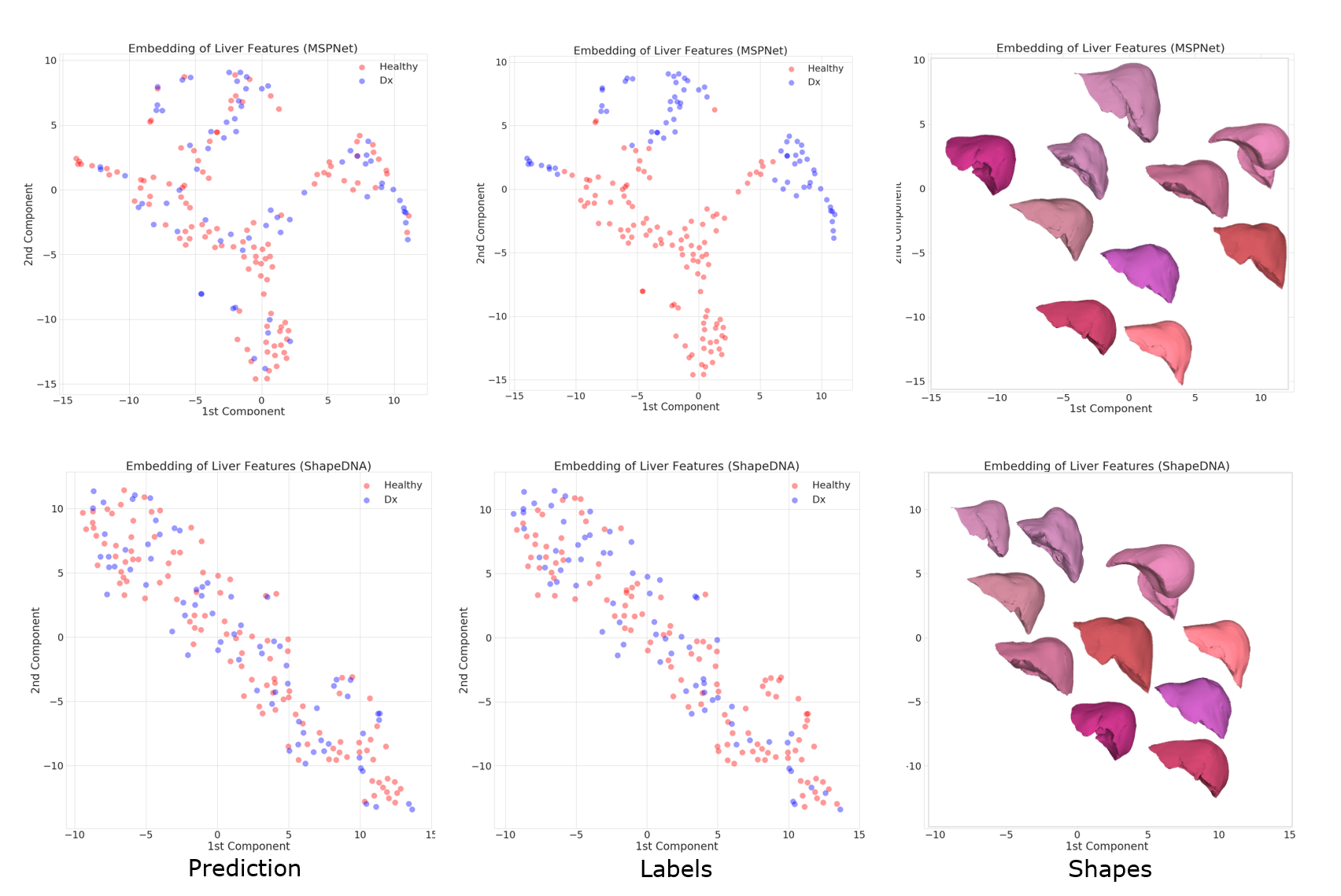}
 		\label{fig:liver_pointnet_embedding}
 	\caption{Embedding of the shape descriptors $\mathbf{S}$ into a 2D space using t-SNE . Each point corresponds to a different subject. The color coding on the left corresponds to the real diagnosis for each subject and the coding in the middle to the label assigned by the classifier. Liver shapes on the right column are shown according to their location in the t-SNE space (the color of the surfaces of the livers is just used for visualization purposes and is not related to the labels).   }\label{fig:embeddings}
 \end{figure*}

One of the main advantages of using representations that are trained in an end-to-end fashion for classification is that the obtained shape representation is specifically optimized for a particular task. In the case of MSPNet, we expect the shape descriptors $\mathbf{S}$ to lie on a space where shapes of organs of healthy patients are clustered close to each other and separated to the shapes of organs of patients diagnosed with diabetes. To have a better understanding of the properties of these learned representations, we visualize 2D projections of the shape descriptors $\mathbf{S}$ by embedding them into a two dimensional space using t-Distributed Stochastic Neighbor Embedding  (t-SNE) \cite{tsne}.  These embeddings can be observed in figure \ref{fig:embeddings}, where we present embeddings on the 2D space for the liver using both AbdomenPrint and MSPNet. In this figure we can observe that the feature space obtained using MSPNet leads to clusters which group  together either healthy subjects or individuals diagnosed with pre-diabetes or diabetes. This can be explained by the fact that the shape descriptors $\mathbf{S}$ of MSPNet are specifically optimized for the separation between these two classes as opposed to AbdomenPrint, which uses standard descriptors that are not targeted to a specific task.

\section{Conclusions}
We have proposed the use of a deep learning based representation for the morphological analysis of abdominal organs, and we have applied this representation for the task of diabetes classifications. Our results show that the use of learning representations based on deep networks have the potential to uncover shape deformations correlated to disease, and potentially to other factors. Compared to other methods, which rely on engineered features, the shape descriptors learned by MSPNet  are optimized for the task of diabetes classification and are based on a simple point cloud representation without the need of calculating meshes or finding points correspondences between subjects.

\subsection{Acknowledgments.} This work was supported in part by DFG,  SAP SE and the Bavarian State Ministry of Education, Science and the Arts in the framework of the Centre Digitalisation.Bavaria (ZD.B).

%\section{BP text}
%We use BrainPrint~\cite{Wachinger2015brainprint} as representation of brain morphology based on the automated segmentation of anatomical brain structures with FreeSurfer~\cite{fischl2002whole}. 
%BrainPrint uses the spectral shape descriptor shapeDNA~\cite{reuter:cad06} to capture  shape information from cortical and subcortical structures. 
% ShapeDNA is computed from the intrinsic geometry of an object by calculating the Laplace-Beltrami spectrum
% \begin{equation}
%\Delta f = - \lambda f .
%\end{equation} 
%The solution consists of eigenvalue~$\lambda_i \in \mathbb{R}$ and eigenfunction~$f_i$ pairs, sorted by eigenvalues, 
%$0 \leq \lambda_1 \leq \lambda_2 \leq \ldots $ %
% The first $l$ non-zero eigenvalues, computed with the finite element method, form the ShapeDNA: $\mathbf{\lambda} = (\lambda_1, \dots, \lambda_l)$. 
% 
%The morphology of each scan~$I$ is described by the concatenation of the spectra of~$\eta$ brain structures 
%$\Lambda = (\mathbf{\lambda}_1, \ldots, \mathbf{\lambda}_{\eta})$, %
%yielding a $D = l \cdot \eta$ dimensional representation. 

%\section{Bibliography}
\bibliography{biblio}{}
\bibliographystyle{splncs03}

\end{document}